\documentclass[journal]{IEEEtran}
\usepackage{graphicx}
\usepackage{amsfonts}

\usepackage[utf8]{inputenc}

\usepackage[english]{babel}
\usepackage{amsmath, amssymb}
\usepackage{float}
\usepackage{makecell}
\usepackage{multirow}
\usepackage{cite}
\usepackage{amsmath,amssymb,amsfonts}
\usepackage{algorithmic}
\usepackage{graphicx}
\usepackage{textcomp}
\usepackage{xcolor}

\usepackage{subfigure}
\usepackage{epstopdf}

\usepackage{graphicx}
\usepackage{comment}
\usepackage{amsmath,amssymb} 
\usepackage{color}
\usepackage{enumerate}

\usepackage{booktabs}
\usepackage{times}
\usepackage{epsfig}
\usepackage{graphicx}
\usepackage{amsmath}
\usepackage{amssymb}
\usepackage{subfigure}
\usepackage{enumerate}
\usepackage{multirow}
\usepackage{caption}

\def\figheight{0.83}

\usepackage{soul}
\soulregister\cite7 
\soulregister\citep7 
\soulregister\citet7 
\soulregister\ref7 
\soulregister\pageref7 

\UseRawInputEncoding

\begin{document}
%
\title{Motion Projection Consistency Based 3D Human Pose Estimation with Virtual Bones from Monocular Videos}
%
%
%

\author{Guangming~Wang,
       Honghao Zeng, Ziliang Wang, Zhe Liu, and Hesheng Wang
        
\thanks{*This work was supported in part by the Natural Science Foundation of China under Grant 62073222, U21A20480 and U1913204, in part by the Science and Technology Commission of Shanghai Municipality under Grant 21511101900, in part by the Open Research Projects of Zhejiang Lab under Grant 2022NB0AB01. The first two authors contributed equally. Corresponding Author: Hesheng Wang.}
\thanks{Guangming Wang, Honghao Zeng, Ziliang Wang, and Hesheng Wang are with Department of Automation, Key Laboratory of System Control and Information Processing of Ministry of Education, Key Laboratory of Marine Intelligent Equipment and System of Ministry of Education, Shanghai Engineering Research Center of Intelligent Control and Management, Shanghai Jiao Tong University, Shanghai 200240, China. (e-mail: wanghesheng@sjtu.edu.cn).

Zhe Liu is with the Department of Computer Science and Technology,
University of Cambridge, Cambridge CB3 0FD, U.K.

}

}

%
%

\markboth{Journal of \LaTeX\ Class Files,~Vol.~14, No.~8, August~2015}%
{Shell \MakeLowercase{\textit{et al.}}: Bare Demo of IEEEtran.cls for IEEE Journals}
%



\maketitle

\begin{abstract}
Real-time 3D human pose estimation is crucial for human-computer interaction. It is cheap and practical to estimate 3D human pose only from monocular video. However, recent bone splicing based 3D human pose estimation method brings about the problem of cumulative error.
In this paper, the concept of virtual bones is proposed to solve such a challenge. The virtual bones are imaginary bones between non-adjacent joints. They do not exist in reality, but they bring new loop constraints for the estimation of 3D human joints. The proposed network in this paper predicts real bones and virtual bones, simultaneously. The final length of real bones is constrained and learned by the loop constructed by the predicted real bones and virtual bones. Besides, the motion constraints of joints in consecutive frames are considered. The consistency between the 2D projected position displacement predicted by the
network and the captured real 2D displacement by the camera is proposed as a new projection consistency loss for the learning of 3D human pose. The experiments on the Human3.6M dataset demonstrate the good performance of the proposed method. Ablation studies demonstrate the effectiveness of the proposed inter-frame projection consistency constraints and intra-frame loop constraints.
\end{abstract}


\IEEEpeerreviewmaketitle

\section{Introduction}
In recent years, increasing attention is attracted to 3D human pose estimation in videos, due to its wide application in the field of action recognition \cite{8372979}, robot learning \cite{9129823,9606554,maplesszhe}, and robot control \cite{7534850,9442367,9546931,9750110}. The state-of-the-art approaches \cite{20183D,2019weakly,2019Lin} are mostly in two steps, 2D joint detection, and 3D pose estimation from the 2D joints. 

The core of these methods is 3D pose estimation based on 2D joints, whose recognized difficulty is depth ambiguity. The information input is limited to the 2D plane while the goal is to predict the joint coordinates in 3D space, so the depth information needs to be compensated with other information, which nowadays is time and spatial information. {Some approaches\cite{20183D,2019Lin,Luvizon_2018_CVPR,2018Propagating} utilize the information of the adjacent frames and\cite{2019weakly} utilizes the information of the multiview. Specifically, Pavllo et al.\cite{20183D} propose an efficient approach for 3D human pose estimation in video based on dilated temporal convolutions on 2D keypoint trajectories. Chen et al.\cite{2019weakly} first use image-skeleton mapping module to obtain 2D skeleton maps from images, and then a view synthesis module is used to predict 3D pose. Lin et al.\cite{2019Lin} propose a deep learning-based framework that utilizes matrix factorization for sequential 3D human pose estimation with the input 2D joint position. Luvizon et al.\cite{Luvizon_2018_CVPR} propose a multitask framework that can estimate the 2D and 3D pose from images and recognize the action from video sequence. Lee et al.\cite{2018Propagating} propose a new long short-term memory (LSTM)-based deep learning architecture named propagating LSTM networks (p-LSTMs) to infer depth. Compared with the method that directly predict 3D human pose from images, these two-stage work add the intermediate variables in the process of the prediction, which means introduce more constraints. However, joints still have a large of degree of freedom, which is most pronounced at the end of the human body. Chen et al.\cite{2020Anatomy} uses consecutive frames to estimate the middle frame’s 3D pose, which is decomposed into the length and direction prediction of bones. However, the method of obtaining joints by the accumulation of bones will accumulate errors.}

To alleviate the accumulated errors, the first contribution in this paper is to improve the bone prediction network by adding the prediction of virtual bones between non-adjacent joints to the bone prediction network. The loop constraint is constructed by real bones and virtual bones to reduce cumulative error and increase the prediction accuracy. Specifically, to avoid the overfitting caused by the limited number of actors in training datasets, each sampled frame is used to predict the corresponding 3D joint position. The bone lengths are calculated from the estimated 3D joint positions. To obtain the real bone lengths in the current frame, a self-attention module is incorporated to weigh the real bone lengths from sampled frames. The virtual bone lengths are calculated from the 3D joint positions in the current frame directly. The ground truth of bone length is used to optimize the self-attention module. The temporal convolutional network in \cite{20183D} is used to predict the direction of all bones. The final positions of joints are derived from the bone length and direction of all bones through a fully connected network. The motivation to add the bones between non-adjacent joints is based on the idea that adding a proper amount of input virtual bones can increase the accuracy of the final joint prediction and reduce the overfitting to a certain extent.

The other contribution is proposing a new projection consistency loss. According to  \cite{wang2020motion}, even if the $l_1$ mean distance between the ground truth and estimated positions is the same, there will be different distribution for joint positions in time dimension. Some researchers \cite{2020Deep,wang2020motion} make kinematics analysis and propose motion loss. Their ideas are all based on the continuity of the displacement of the joint, which is constrained by the real displacement in 3D. A new projection consistency loss is proposed in this paper, comparing the 2D projection of the 3D displacement of the estimated joints between adjacent frames and the 2D displacement derived from the input 2D keypoints. This loss can reduce the fluctuation of joint position estimation results in the continuous video, with no need of 3D ground truth of joints.

In summary, the approach proposed in this paper makes the following contributions: 
\begin{itemize}
	\vspace{-0pt}
	\item A bone length prediction network with additional bones among non-adjacent joints is presented to avoid overfitting and predict joints more accurately.
	\vspace{-0pt}
	\item A new projection loss based on the 2D displacement of 3D joints is proposed, not only smoothing the error between adjacent frames but also improving the accuracy of 3D joint position prediction.
	\vspace{-0pt}
	\item A variety of virtual bone combination modes are validated. Ablation studies demonstrate the effectiveness of the proposed method. The combination of virtual bones and projection loss lead to a good performance on Human3.6M dataset \cite{Catalin2014Human3}.
	\vspace{-0pt}
\end{itemize}

\begin{figure}
	\centering
	\includegraphics[width = 0.98\columnwidth]{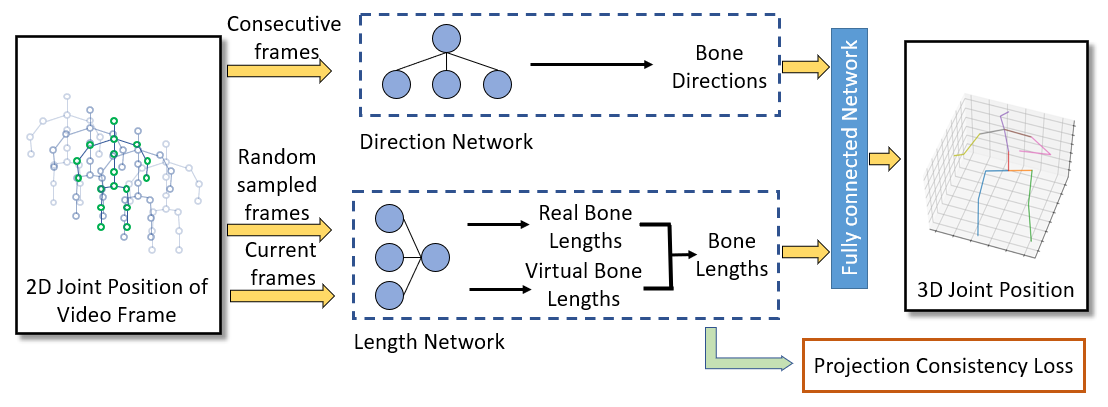}
	\vspace{-2pt}
	\caption{The overview of the proposed framework. It uses consecutive frames to predict the direction of real and virtual bones, randomly sampled frames to predict the length of real bones, and current frames to predict the length of virtual bones. The 3D joint position is predicted by a fully connected network with predicted bones. Meanwhile, a projection consistency loss is used to constrain the learning of bone prediction.}
	\label{fig:overview}
	\vspace{-0pt}
\end{figure}

\section{Related Work}
The development of the field of 3D human pose estimation has undergone a variety of method changes. One of them uses neural networks to estimate the 3D pose of the human body directly from the input image, including  \cite{2019Multiview,2020Moulding,2020VIBE}. Mitra et al. \cite{2019Multiview} first train the neural network in an unsupervised fashion with multi-view data to get the human pose representation. Then, ground truth is used to map this representation to the 3D human joints. It belongs to a semi-supervised method. Gabeur et al.~\cite{2020Moulding} use the structure of encoder-decoder to estimate the depth of the front and rear surfaces of the human body as the representation of 3D pose. Considering the time information, Kocabas et al.~\cite{2020VIBE} use the video as the input, obtaining 82 parameters of the Skinned Multi-Person Linear model (SMPL)~\cite{loper2015smpl} to represent the shape and pose of the human body.

Recently, the task of 3D human joint estimation is divided into two parts. 2D joint positions are first obtained from images, and then 3D joint positions are predicted from the 2D position.
Methods for detecting 2D joint positions from images have become mature~\cite{2019Zhang,2019Single}. Zhang et al.~\cite{2019Zhang} optimize the probability distribution obtained from the heat map, which leads to better 2D joint estimation results. Nie et al. \cite{2019Single} propose a chain method to represent joint coordinates. Each joint coordinate is represented by the root joint coordinate and the bone vector between adjacent joints.

With the help of mature technology for estimating 2D keypoints, researchers estimate 3D poses on this basis, such as \cite{wang20203d,2020Weakly,li2020cascaded,20183D,2020Attention,20203DCheng,20203DWu,2019Reweighted,2020Anatomy,2018Propagating}. Lee et al.~\cite{2018Propagating} firstly use convolutional neural network to extract a 2D pose from RGB image, and then they propose the p-LSTMs model to infer depth to obtain 3D human pose.
Wang et al.~\cite{wang20203d} notice the importance of the occlusion relationship of the joints, which is used as part of the loss for the network training. 
Iqbal et al.~\cite{2020Weakly} first extract 2D joint positions from the input multi-view images, and then estimate 3D positions from the 2D positions and the camera external parameters through an optimization method. An unsupervised loss is constructed by the consistency of the poses from different viewing angles. Different from other works, Li et al.~\cite{li2020cascaded} estimate the length and direction angle of each bone and propose a dataset augmentation method that improves the accuracy of the algorithm in unusual poses.

Recent studies \cite{20183D,2020Attention,20203DCheng,20203DWu,2019Reweighted} take into account the time information and use video as input instead of a separate frame. Pavllo et al.~\cite{20183D} regard the process of estimating 3D position from 2D as the encoding part. The process of projecting 3D back to 2D is regarded as the decoding part. They train with the labeled data and use unlabeled data to calculate the consistency of input and decoded output as an unsupervised loss. Liu et al.~\cite{2020Attention} introduce an attention mechanism. Different frames are weighed and different convolution kernels are used to improve network structure and algorithm performance. Cheng et al.~\cite{20203DCheng} apply the method of multi-scale analysis in the time and space dimensions to deal with the problems of different sizes and different speeds of humans. Chen et al.~\cite{2020Anatomy} decompose the 3D pose into the bone length and the bone direction. Considering the length invariance of bones and the visibility of the joints on the image, they use the whole video combined with the attention mechanism to obtain a more accurate bone length estimation. At the same time, a new layered bone direction prediction network is proposed to get better results. Wu et al.~\cite{20203DWu} establish a depth map of joints to calculate the limb depth map. Then, the hidden information extracted from the picture is combined to directly obtain 3D poses. Jiang et al.~\cite{2019Reweighted} pretrain a series of 3D poses using input pictures labeled with 2D joint information. They take the weights of these poses from the model to get a rough pose and use the residual compensation to obtain the final predicted pose.
To increase the accuracy, some researchers use the motion of the human joints in the video to improve the method of supervision. For example, Xu et al.~\cite{2020Deep} take the kinematics analysis for monocular 3D human pose estimation between multiple frames to correct the limbs at the end of the human body.

\begin{figure}
	\centering
	\subfigure[]{\label{fig2:subfiga}\includegraphics[width=0.23\textwidth]{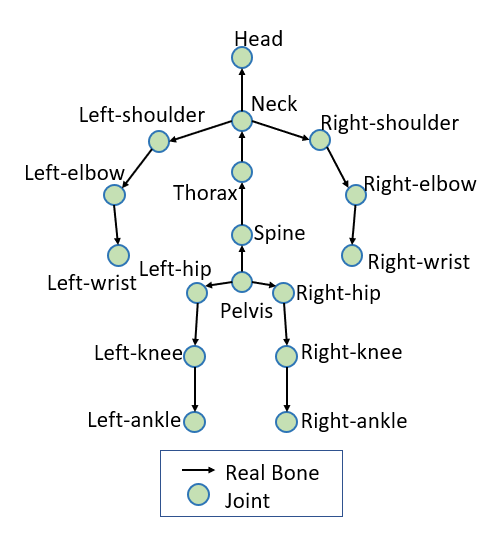}}
	\subfigure[]{\label{fig2:subfigb}\includegraphics[width=0.23\textwidth]{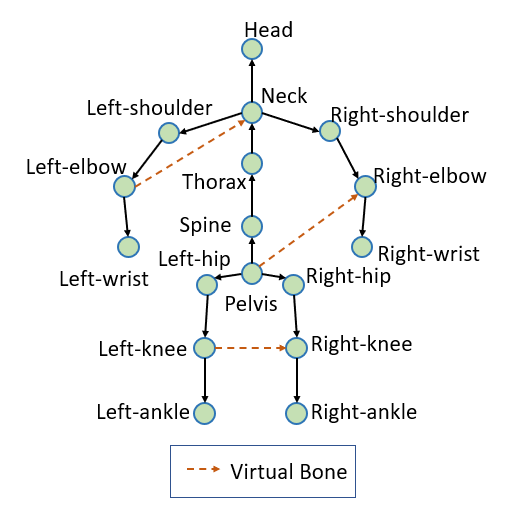}}
	\vspace{-3pt}
	\caption{(a) The common skeleton of human with joints and bones for representation. (b) Schematic diagram of virtual bones, such as left-elbow to neck, pelvis to right-elbow, and left-knee to right-knee.}
	\vspace{-0pt}
\end{figure}

Meanwhile, some researchers use distance matrix to measure the relationship between different joints, such as \cite{moreno20173d,guo2018occluded,gao2019optimized}. Noguer~\cite{moreno20173d} proposes the Euclidean
Distance Matrice (EDM) between 2D keypoints to estimate the EDM between 3D human joints, finally leading to the human pose. Guo et al.~\cite{guo2018occluded} based on \cite{moreno20173d}, recover the occluded joints before estimating 3D EDM. Gao et al.~\cite{gao2019optimized} noticed the distance relationship between non-adjacent joints when determining the distance matrix.

\section{Motion Projection Consistency Based 3D Human Pose Estimation with Virtual Bones}

The overview of the proposed method is illustrated in Figure \ref{fig:overview}. In this section, we elaborate on the details of our method. In Section \ref{subsection:3.1}, we introduce the architecture of the bone prediction network with the virtual bone output. In Section \ref{subsection:3.2}, we introduce the detail of obtaining 3D joint position from the bones. In Section \ref{subsection:3.3}, we present the detail of projection consistency loss. In Section \ref{subsection:3.4}, the other used losses are introduced.
\subsection{Bone Length and Direction Prediction Networks}
\label{subsection:3.1}

In recent years, the human skeleton structure commonly used by researchers is shown in Figure \ref{fig2:subfiga}, in which the bones between adjacent joints are recorded as real bones. In this paper, we use the concept of virtual bones to represent the imaginary bones between non-adjacent joints, as in Figure \ref{fig2:subfigb}. These two terms will be used frequently in the following sections. In this paper, the learning of real bones is optimized through the prediction of virtual bones and real bones simultaneously.

\begin{figure*}[t]
	\centering
	\includegraphics[width=0.95\textwidth]{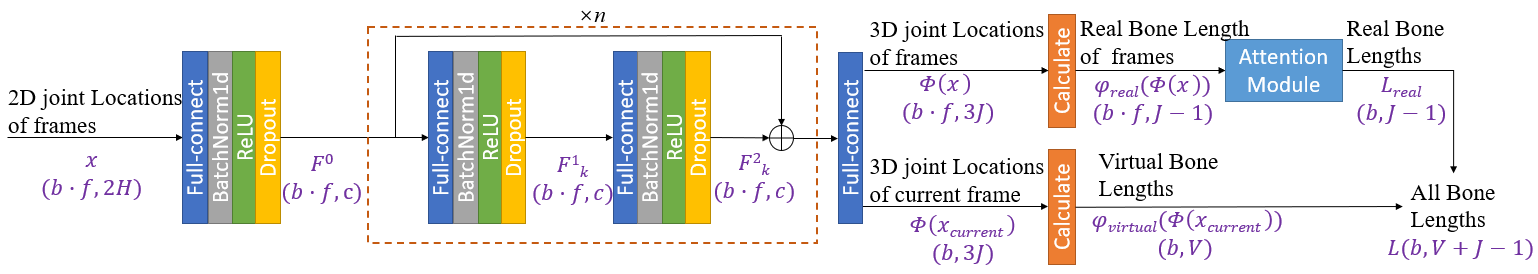}
	\vspace{-3pt}
	\caption{The detailed structure of bone length prediction network.  {The input of this network is the 2D joint location $x$ in random $f$ frames from a video. $F^0$ is intermediate features. $n$ is the number of residual blocks. $F_{k}^1$ and $F_{k}^2$ are intermediate features in the $k$-th residual block. $x_{current}$ is the 2D joint location in the current frame. The 3D joint location of all frames $\phi(x)$ and 3D joint locations of current frame $\phi(x_{current})$ are obtained through the coarse joint location prediction network $\phi$. Then, real bone length $\varphi_{real}(\phi(x))$ and virtual bone length $\varphi_{virtual}(\phi(x_{current}))$ are obtained by calculating. Final real bones $L_{real}$ in the current frame are obtained with attention module. $b$ is the batchsize. $c$ is the number of feature dimension.}}

	\label{fig:3}
	\vspace{-0pt}
\end{figure*}

The structure of the bone length prediction network is shown in Figure \ref{fig:3}. The bone length prediction network not only predicts the bone length of real bones but also predicts the bone length of virtual bones. To get global information more effectively, like \cite{2020Anatomy}, {we input 2D joint location $x$ of $J$ joints from $f$ frames sampled from a video to the network. The 2D joint locations are first used to predict coarse 3D locations of the $J$ joints, which are utilized to calculate the length of bones.} 

Because of the invariance of the bone length of real bones, the length of real bones is predicted for each of the random frames and weighed by an attention mechanism to get the length of real bones in the current frame:
\begin{equation}
	L_{real} = \sum_{i=1}^{f} w_i \varphi_{real}(\phi(x_i)),
\end{equation}
where $w_i$ represents the matrix composed of the weight of each bone in $i$-th frame. $x_i$  $(i=1, 2,\dots,f)$ represents the input 2D joint location in random frames. $\varphi_{real}$ represents the calculation to obtain the length of the real bones from coarse 3D joint location in frames. $\phi$ represents the network that predicts the coarse 3D joint location from random frames.

However, the bone length of virtual bone varies from frame to frame, so only the current frame is used, skipping the attention mechanism, {to predict the bone length of $V$ virtual bones}:
\begin{equation}
	L_{virtual} = \varphi_{virtual}(\phi(x_{current})),
\end{equation}
where $x_{current}$ represents the 2D joint location in frame at the current timestamp. $\varphi_{virtual}$ represents the calculation to obtain the length of virtual bones from coarse 3D joint location in current frame. $\phi$ represents the network that predicts coarse 3D joint location from the current frame.

The method for predicting bone direction is the same as \cite{20183D}, utilizing the temporal fully-convolutional network whose output is the unit vector of bone direction:
\begin{equation}
	D_{o}=\psi(x_1, x_2, \dots, x_f),
\end{equation}
where $x_k$  $(k=1, 2,\dots,f)$ represents the input 2D joint location in consecutive frames. $\psi$ represents the temporal fully-convolutional network.

\begin{figure*}
	\centering
	\vspace{-0pt}
	\subfigure[5 virtual bones]{\label{fig4:subfiga}\includegraphics[width=0.23\textwidth]{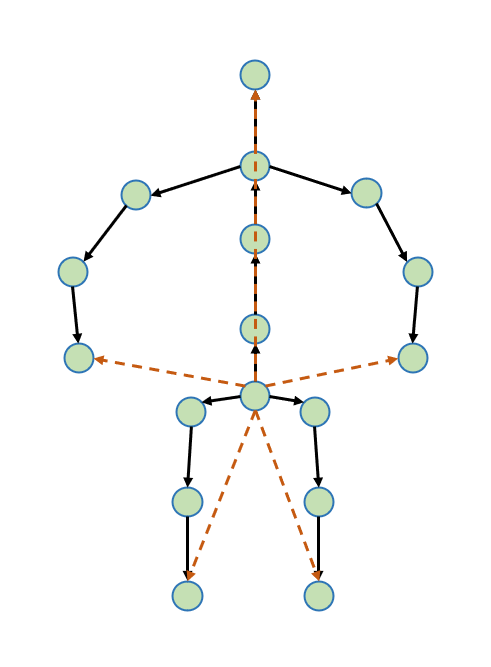}}
	\subfigure[10 virtual bones]{\label{fig4:subfigb}\includegraphics[width=0.23\textwidth]{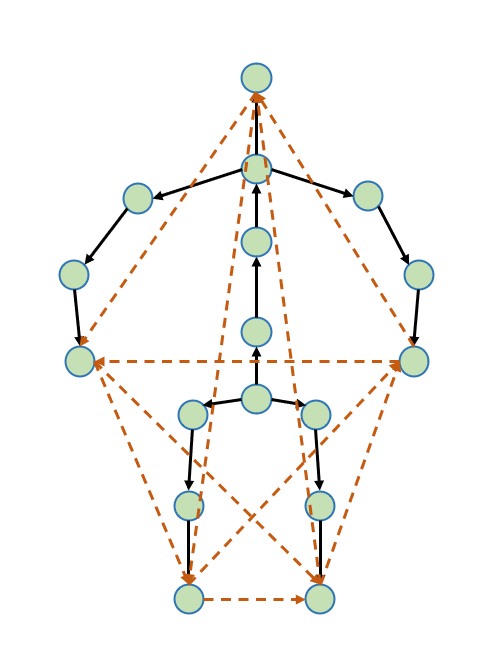}}
	\subfigure[13 virtual bones]{\label{fig4:subfigc}\includegraphics[width=0.23\textwidth]{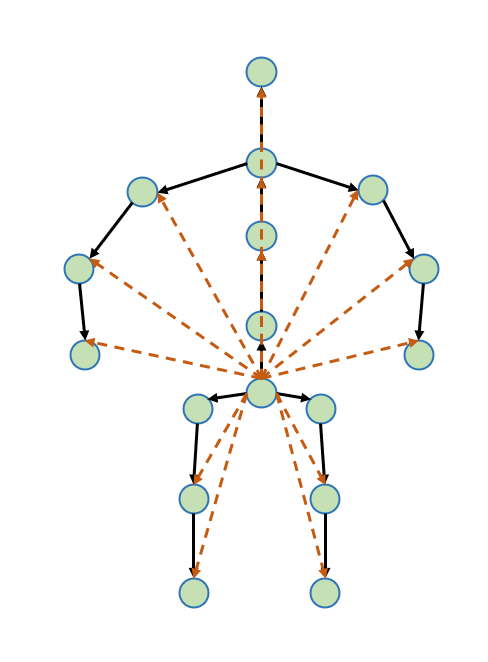}}
	\subfigure[23 virtual bones]{\label{fig4:subfigd}\includegraphics[width=0.23\textwidth]{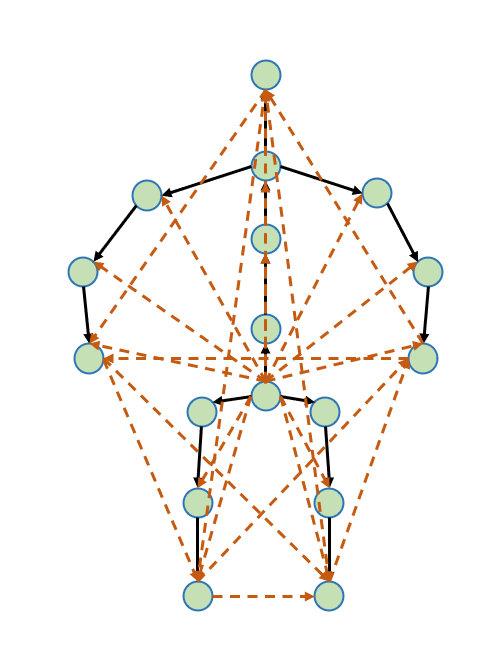}}
	\vspace{-3pt}
	\caption{Schematic diagram of virtual bones added to the network. (a) 5 bones from pelvis to the joints at the end of the human body. (b) 10 bones between the joints at the end of the human body. (c) 13 bones from pelvis to every non-adjacent joint. (d) 23 bones, the combination of (b) and (c).}
	\vspace{-0pt}
	\label{fig:fb}
\end{figure*}

{For the standard 17-joint human skeleton structure, there is only one path from the root joint to the target joint. What our network predicts are the bone length and unit orientation of the bone, so with the bone vectors on the path from root joint to the target joint, we only get the unique target joint coordinates. After the introduction of virtual bones, for a target joint, we will get multiple joint coordinates along different paths. The final target joint coordinate is obtained by weighting multiple predicted values, which can reduce the accumulate error and increase the prediction accuracy of target joint.}

The joints at the end of the human body are significantly more unstable than others\cite{2020Deep}, so the virtual bones input the network are selected. Moreover, considering the human skeleton is based on root joint (i.e. pelvis), we select the bones related to the four joints at the end of the human body (i.e. Head, Left-wrist, Left-ankle, Right-wrist, Right-ankle) and the root joint of the human body. 

Finally, 4 options are determined as shown in Figure \ref{fig:fb}. Option one, as shown in Figure \ref{fig4:subfiga}, contains 5 virtual bones between the root joint and joints at the end of the human body.  Option two, as shown in Figure \ref{fig4:subfigb}, contains 10 virtual bones among joints at the end of the human body. Option three, as shown in Figure \ref{fig4:subfigc}, contains 13 virtual bones between the root joint and other non-adjacent joints. Option four, as shown in Figure \ref{fig4:subfigd}, contains 23 virtual bones mentioned in Figure \ref{fig4:subfigb} and Figure \ref{fig4:subfigc}. Experiments are conducted with these options separately to compare their performance. The experiment detail is in Section \ref{subsec:ablation}.

\subsection{3D Joint Prediction}
\label{subsection:3.2}
Usually the $k$-th joint's position, $P_k$ is derived as:
\begin{equation}
	P_k = \sum_{m\in R_k}D_{o,m}\cdot L_{m},
	\label{eq:joint}
\end{equation}
where $D_{o,m}\in D_o$ and $L_m\in L=L_{real}\cup L_{virtual}$ are the direction and length of bone $m$. $R_k$ is the collection of all bones along the path of the normal human skeleton from the root joint (i.e. pelvis) to the $k$-th joint.

However, the equation~(\ref{eq:joint}) only considers the path along with the real bones. If the virtual bones are added, there will be more than one path to a joint. Hence the equation changes to:
\begin{equation}
	P_k = \sum_{R_{k, i}\in \Lambda_k }\omega_{k,i}\sum_{m\in R_{k,i}}D_{o,m}\cdot L_{m},
\end{equation}
where $\Lambda_k$ represents the set of all the paths from the root joint to the $k$-th joint. $w_{k,i}$ is the weight of the $i$-th path. $R_{k,i}$ is the collection of all the bones on the $i$-th path to the $k$-th joint.

Thus, in the proposed approach, there are different ways to obtain the position of a joint. A fully connected network is used to calculate the position of joints with the direction and length of both real and virtual bones as input. The network will automatically adjust the weight distribution of every bone related to every predicted joint to obtain the 3D position. The way obtaining the joints is determined by the bones input, so the selected virtual bones are key of the proposed approach. At the beginning of the experiment, we try the all real bones and virtual bones that between every joints (e.g. 17-joint-skeleton has 136 bones). However, because the number of bones to be predicted is too large, the time required for each training epoch is too long. In addition, the experimental effect is not good, so we gave up using all the bones.

\begin{figure}
	\centering
	\includegraphics[width=0.45\textwidth]{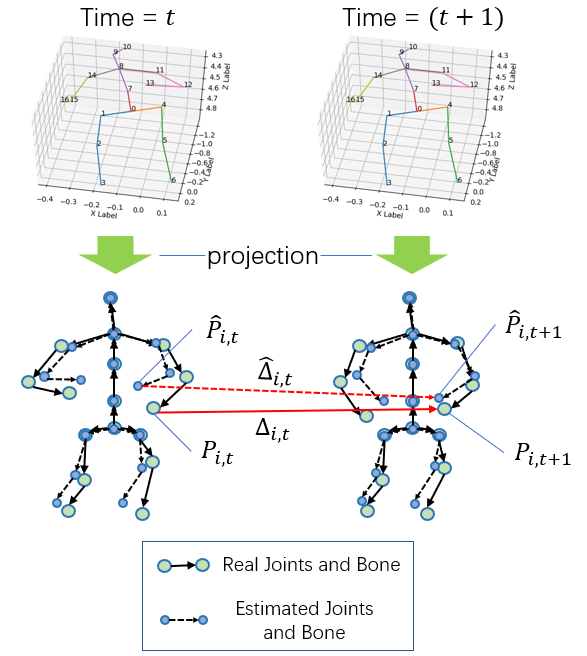}
	\vspace{-0pt}
	\caption{Schematic diagram of the projection consistency loss. In the figure, $\hat P_{i,t}$ represents the estimated projection position of the $i$-th joint in the $t$-th frame.  $P_{i,t}$ is the ground truth projection of $i$-th joint position in the $t$-th frame. Similarly, $\hat P_{i,t+1}$ represents the estimated projection position of the $i$-th joint in the $(t+1)$-th frame. $P_{i,t+1}$ is the ground truth projection of the $i$-th joint position in the $(t+1)$-th frame. $\hat{\Delta}_{i,t}$ means the 2D displacement of the $i$-th joint's estimated projection at time $t$. $\Delta_{i,t}$ means the 2D displacement of the $i$-th joint's ground truth projection at time $t$.} 
	\vspace{-0pt}
	\label{fig:proj method}
\end{figure}

\subsection{Projection Consistency Loss}
\label{subsection:3.3}

In this section, a new loss function, projection consistency loss, is designed. As shown in Figure \ref{fig:proj method}, considering the position of the 2D joints in two adjacent frames captured by the same camera, each joint has a certain displacement. Naturally, the 2D projection of the estimated 3D joint position will move in the same way as the 2D input. The calculation process is as follows.

First, according to the pinhole camera model, the estimated 3D position of each joint is projected back to the 2D plane:
\begin{equation}
	\hat{Z_c}\left(
	\begin{matrix}
		\hat u\\
		\hat v\\
		1
	\end{matrix}\right)
	=\left(
	\begin{matrix}
		\frac{f}{d_x} & 0 & u_0 \\
		0 &\frac{f}{d_y} & v_0 \\
		0 & 0 & 1
	\end{matrix}\right)\left(
	\begin{matrix}
		\hat{X_c}\\
		\hat{Y_c}\\
		\hat{Z_c}
	\end{matrix}
	\right),
\end{equation}
where $f$ represents the focal length of the camera. $(u_0, v_0)$ represents the position of the optical center of the camera. $(\hat{X_c}, \hat{Y_c}, \hat{Z_c})$ is the estimated joint coordinates in the camera coordinate system. $(\hat{u}, \hat{v})$ represents the coordinates of the estimated 2D projection of the joint. $d_x$, $d_y$ represent zoom factors.

\begin{figure}[t]
	\centering
	\includegraphics[width=0.45\textwidth]{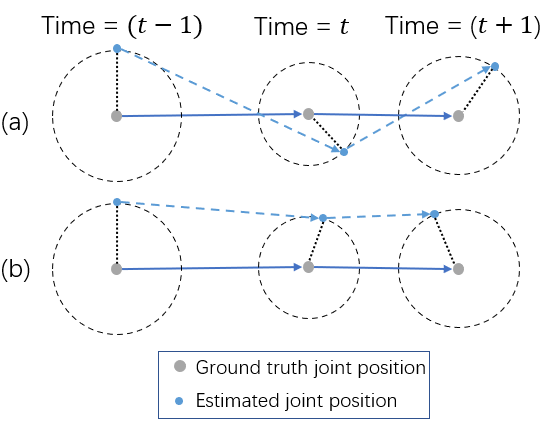}
	\vspace{-0pt}
	\caption{Comparison of projection consistency loss and projection loss, where (a) and (b) are two possible trajectories of the 2D projection of a certain joint's estimated position. The single frame projection loss is exactly the same in (a) and (b), while the projection consistency loss is completely different.} 
	\label{fig:proj benefit}
	\vspace{-0pt}
\end{figure}

Secondly, the 2D displacement of the estimated joint projection is calculated as:
\begin{equation}
	\hat{\Delta}_{i,t} = \left(
	\begin{matrix}
		\hat{u}_{i,t}\\
		\hat{v}_{i,t}
	\end{matrix}
	\right) - \left(
	\begin{matrix}
		\hat{u}_{i,t-1}\\
		\hat{v}_{i,t-1}
	\end{matrix}
	\right),
\end{equation}
where $(\hat{u}_{i,t}, \hat{v}_{i,t})$ represents the $i$-th joint's estimated 2D projection at time $t$. $\hat{\Delta}_{i,t}$ means the estimated 2D displacement of the $i$-th joint's projection from time $t-1$ to time $t$.\par 
Similarly, the ground truth 2D displacement is calculated as:
\begin{equation}
	\Delta_{i,t} = \left(
	\begin{matrix}
		u_{i,t}\\
		v_{i,t}
	\end{matrix}
	\right) - \left(
	\begin{matrix}
		u_{i,t-1}\\
		v_{i,t-1}
	\end{matrix}
	\right),
\end{equation}
where $(u_{i,t}, v_{i,t})$ represents the $i$-th joint's ground truth 2D projection at time $t$. $\Delta_{i,t}$ means the 2D displacement of the $i$-th joint's ground truth projection from time $t-1$ to time $t$.

\begin{table*}[t]
\subtable[Comparison under the metric Protocol 1]{
\resizebox{\textwidth}{!}{
		\begin{tabular}{|ccccccccccccccccc|}
			\hline
			Methods                    & Dir. & Disc. & Eat  & Greet & Phone & Photo & Pose & Purch. & Sit  & SitD. & Smoke & Wait & WalkD. & Walk & WalkT. & Avg  \\
			Martinez et al.  \cite{Martinez_2017_ICCV} ICCV’17     & 51.8          & 56.2          & 58.1          & 59.0            & 69.5          & 78.4          & 55.2          & 58.1          & 74.0            & 94.6          & 62.3          & 59.1          & 65.1          & 49.5          & 52.4          & 62.9          \\
			Sun et al.  \cite{Sun_2017_ICCV} ICCV17         & 52.8          & 54.8          & 54.2          & 54.3          & 61.8          & 67.2          & 53.1          & 53.6          & 71.7          & 86.7          & 61.5          & 53.4          & 61.6          & 47.1          & 53.4          & 59.1          \\
			Pavlakos et al.  \cite{Pavlakos_2018_CVPR} CVPR18    & 48.5          & 54.4          & 54.4          & 52.0            & 59.4          & 65.3          & 49.9          & 52.9          & 65.8          & 71.1          & 56.6          & 52.9          & 60.9          & 44.7          & 47.8          & 56.2          \\
			Yang et al.  \cite{Yang_2018_CVPR} CVPR18        & 51.5          & 58.9          & 50.4          & 57.0            & 62.1          & 65.4          & 49.8          & 52.7          & 69.2          & 85.2          & 57.4          & 58.4          & 43.6          & 60.1          & 47.7          & 58.6          \\
			Luvizon et al.  \cite{Luvizon_2018_CVPR} CVPR18     & 49.2          & 51.6          & 47.6          & 50.5          & 51.8          & 60.3          & 48.5          & 51.7          & 61.5          & 70.9          & 53.7          & 48.9          & 57.9          & 44.4          & 48.9          & 53.2          \\
			Hossain \& Little  \cite{hossain2018exploiting} ECCV18  & 48.4          & 50.7          & 57.2          & 55.2          & 63.1          & 72.6          & 53.0            & 51.7          & 66.1          & 80.9          & 59.0            & 57.3          & 62.4          & 46.6          & 49.6          & 58.3          \\
			Lee et al.  \cite{2018Propagating} ECCV18         & 40.2 & 49.2          & 47.8          & 52.6          & 50.1          & 75.0            & 50.2          & 43.0            & 55.8          & 73.9          & 54.1          & 55.6          & 58.2          & 43.3          & 43.3          & 52.8          \\
			Chen et al.  \cite{2019weakly} CVPR’19           & 41.1 & 44.2  & 44.9 & 45.9  & 46.5  & 39.3  & 41.6 & 54.8   & 73.2 & 46.2  & 48.7  & 42.1 & 35.8   & 46.6 & 38.5   & 46.3 \\
			Pavllo et al. \cite{20183D} (243 frames, Causal) CVPR’19 & 45.9          & 48.5          & 44.3          & 47.8          & 51.9          & 57.8          & 46.2          & 45.6          & 59.9          & 68.5          & 50.6          & 46.4          & 51.0          & 34.5          & 35.4          & 49.0          \\
			Pavllo et al.  \cite{20183D} (243  frames) CVPR’19         & 45.2 & 46.7  & 43.3 & 45.6  & 48.1  & 55.1  & 44.6 & 44.3   & 57.3 & 65.8  & 47.1  & 44.0 & 49.0   & 32.8 & 33.9   & 46.8 \\
			Lin et al.  \cite{2019Lin} BMVC’19            & 42.5 & 44.8  & 42.6 & 44.2  & 48.5  & 57.1  & 42.6 & 41.4   & 56.5 & 64.5  & 47.4  & 43.0 & 48.1   & 33.0 & 35.1   & 46.6 \\
			Cai et al.  \cite{9009459} ICCV’19            & 44.6 & 47.4  & 45.6 & 48.8  & 50.8  & 59.0  & 47.2 & 43.9   & 57.9 & 61.9  & 49.7  & 46.6 & 51.3   & 37.1 & 39.4   & 48.8 \\
			Yeh et al.  \cite{2019Yeh} NIPS’19            & 44.8 & 46.1  & 43.3 & 46.4  & 49.0  & 55.2  & 44.6 & 44.0   & 58.3 & 62.7  & 47.1  & 43.9 & 48.6   & 32.7 & 33.3   & 46.7 \\
			Xu et al. \cite{2020Deep} CVPR’20                              & 37.4          & 43.5          & 42.7          & 42.7          & 46.6          & 59.7          & 41.3          & 45.1          & 52.7          & 60.2          & 45.8          & 43.1          & 47.7          & 33.7          & 37.1          & 45.6          \\
			Chen et al.  \cite{2020Anatomy} (9 frames) IEEE T-CSVT'2021 & 44.4 & 47.0  & 43.8 & 46.2  & 49.6  & 57.1  & 46.0 & 44.0   & 55.9 & 61.1  & 48.5  & 45.0 & 49.4   & 35.6 & 39.0   & 47.5 \\ 
			Chen et al. \cite{2020Anatomy} (243 frames) IEEE T-CSVT'2021 & 41.4          & 43.5          & 40.1          & 42.9          & 46.6          & 51.9          & 41.7          & 42.3          & 53.9          & 60.2          & 45.4          & 41.7          & 46.0          & 31.5          & 32.7          & 44.1          \\
			\hline
			ours (9 frames)    & 44.6          & 46.1          & 44.9          & 46.2          & 49.5          & 57.1          & 45.2          & 43.4          & 55.6         & 61.2          & 48.2          & 44.7          & 49.1          & 25.6          & 39.3          & 47.4          \\
			ours (243 frames)                              & 42.4          & 43.5          & 41.0          & 43.5          & 46.7          & 54.6          & 42.5          & 42.1          & 54.9          & 60.5          & 45.7          & 42.1          & 46.5          & 31.7          & 33.7          & 44.8          \\
			\hline
			\end{tabular}}
			}
			\qquad
			\subtable[Comparison under the metric Protocol 2]{
			\resizebox{\textwidth}{!}{
			\begin{tabular}{|ccccccccccccccccc|}
			\hline
			Methods                    & Dir. & Disc. & Eat  & Greet & Phone & Photo & Pose & Purch. & Sit  & SitD. & Smoke & Wait & WalkD. & Walk & WalkT. & Avg  \\
			Martinez et al.  \cite{Martinez_2017_ICCV} ICCV’17   & 39.5          & 43.2          & 46.4          & 47.0            & 51.0            & 56.0            & 41.4          & 40.6          & 56.5          & 69.4          & 49.2          & 45.0            & 49.5          & 38.0            & 43.1          & 47.7          \\
			Sun et al.  \cite{Sun_2017_ICCV} ICCV’17        & 42.1          & 44.3          & 45.0            & 45.4          & 51.5          & 53.0            & 43.2          & 41.3          & 59.3          & 73.3          & 51.0            & 44.0            & 48.0            & 38.3          & 44.8          & 48.3          \\
			Pavlakos et al.  \cite{Pavlakos_2018_CVPR} CVPR’18   & 34.7          & 39.8          & 41.8          & 38.6          & 42.5          & 47.5          & 38.0            & 36.6          & 50.7          & 56.8          & 42.6          & 39.6          & 43.9          & 32.1          & 36.5          & 41.8          \\
			Yang et al.  \cite{Yang_2018_CVPR} CVPR18        & 26.9          & 30.9          & 36.3          & 39.9          & 43.9          & 47.4          & 28.8          & 29.4          & 36.9          & 58.4          & 41.5          & 30.5          & 29.5          & 42.5          & 32.2          & 37.7          \\
			Hossain \& Little  \cite{hossain2018exploiting} ECCV’18 & 35.7          & 39.3          & 44.6          & 43            & 47.2          & 54            & 38.3          & 37.5          & 51.6          & 61.3          & 46.5          & 41.4          & 47.3          & 34.2          & 39.4          & 44.1          \\
			Chen et al.  \cite{2019weakly} CVPR’19           & 36.9 & 39.3  & 40.5 & 41.2  & 42.0  & 34.9  & 38.0 & 51.2   & 67.5 & 42.1  & 42.5  & 37.5 & 30.6   & 40.2 & 34.2   & 41.6 \\
			Pavllo et al.  \cite{20183D} (243 frames, Causal) CVPR’19 & 35.1          & 37.7          & 36.1          & 38.8          & 38.5          & 44.7          & 35.4          & 34.7          & 46.7          & 53.9          & 39.6          & 35.4          & 39.4          & 27.3          & 28.6          & 38.1          \\
			Pavllo et al.   \cite{20183D} (243 frames) CVPR’19         & 34.1 & 36.1  & 34.4 & 37.2  & 36.4  & 42.2  & 34.4 & 33.6   & 45.0 & 52.5  & 37.4  & 33.8 & 37.8   & 25.6 & 27.3   & 36.5 \\
			Lin et al.  \cite{2019Lin} BMVC’19            & 32.5 & 35.3  & 34.3 & 36.2  & 37.8  & 43.0  & 33.0 & 32.2   & 45.7 & 51.8  & 38.4  & 32.8 & 37.5   & 25.8 & 28.9   & 36.8 \\
			Cai et al.  \cite{9009459} ICCV’19            & 35.7 & 37.8  & 36.9 & 40.7  & 39.6  & 45.2  & 37.4 & 34.5   & 46.9 & 50.1  & 40.5  & 36.1 & 41.0   & 29.6 & 33.2   & 39.0 \\
			Xu et al. \cite{2020Deep} CVPR’20                              & 31.0          & 34.8          & 34.7          & 34.4          & 36.2          & 43.9          & 31.6          & 33.5          & 42.3          & 49.0          & 37.1          & 33.0          & 39.1          & 26.9          & 31.9          & 36.2          \\
			Chen et al.  \cite{2020Anatomy} (9 frames) IEEE T-CSVT'2021 & 33.6 & 36.8  & 34.6 & 37.4  & 37.8  & 43.3  & 34.6 & 33.6   & 43.9 & 49.5  & 38.8  & 34.5 & 38.9   & 27.5 & 31.6   & 37.1 \\ 
			Chen et al. \cite{2020Anatomy} (243 frames) IEEE T-CSVT'2021 & 32.6          & 35.1          & 32.8          & 35.4          & 36.3          & 40.4          & 32.4          & 32.3          & 42.7          & 49.0          & 36.8          & 32.4          & 36.0          & 24.9          & 26.5          & 35.0          \\ \hline
			Ours (9 frames)      & 33.3          & 36.3          & 35.1          & 37.0          & 37.5          & 42.9          & 34.1          & 33.0          & 43.6         & 49.3          & 38.3          & 34.3          & 38.7          & 27.4          & 31.7          & 36.8          \\ 
			Ours (243 frames)                              & 32.2          & 34.9          & 33.0          & 35.2          & 35.7          & 40.7          & 32.6          & 32.1          & 42.8          & 48.9          & 36.5          & 32.5          & 35.9          & 25.0          & 26.7          & 34.9          \\
			\hline
			\end{tabular}}
			}
			\qquad
			\subtable[Comparison under the metric MPJVE]{
			\resizebox{\textwidth}{!}{
			\begin{tabular}{|ccccccccccccccccc|}
			\hline
			Methods                         & Dir. & Disc. & Eat  & Greet & Phone & Photo & Pose & Purch. & Sit  & SitD. & Smoke & Wait & WalkD. & Walk & WalkT. & Avg  \\
			Pavllo et al.  \cite{20183D} (243 frames) CVPR’19         & 3.0  & 3.1   & 2.2  & 3.4   & 2.3   & 2.7   & 2.7  & 3.1    & 2.1  & 2.9   & 2.3   & 2.4  & 3.7    & 3.1  & 2.8    & 2.8  \\
			Chen et al.  \cite{2020Anatomy} (9 frames) IEEE T-CSVT'2021 & 4.2  & 4.2   & 3.3  & 4.7   & 3.5   & 3.8   & 3.8  & 3.9    & 3.3  & 4.0   & 3.5   & 3.6  & 4.7    & 4.2  & 4.1    & 3.9  \\ 
			Chen et al. \cite{2020Anatomy} (243 frames) IEEE T-CSVT'2021 & 2.7           & 2.8           & 2.0           & 3.1           & 2.0           & 2.4           & 2.4           & 2.8           & 1.8           & 2.4           & 2.0           & 2.1           & 3.4           & 2.7           & 2.4           & 2.5           \\
			\hline
			Ours (9 frames)       & 4.2           & 4.2           & 3.3           & 4.7           & 3.5           & 3.8           & 3.8           & 3.9           & 3.2          & 3.9           & 3.5           & 3.6           & 4.8           & 4.3           & 4.1           & 3.9           \\
			Ours (243 frames)                              & 2.7           & 2.8           & 2.0           & 3.1           & 2.0           & 2.4           & 2.4           & 2.7           & 1.8           & 2.4           & 2.0           & 2.1           & 3.3           & 2.7           & 2.4           & 2.5           \\
			\hline
		\end{tabular}
		}
		}
	\vspace{-0pt}
	\caption{Comparisons of the proposed method with other existing methods in all actions of Human3.6M dataset under the metrics Protocol 1, Protocol 2, and MPJVE. The results are based on 2D joint input from CPN \cite{20183D}.}
	\label{table:ex}
	\vspace{-0pt}
\end{table*}

Finally, the projection consistency loss function is given as follows:
\begin{equation}
	Loss_{\text{proj}} = \text{mean}(\left\| \hat{\Delta}_{i,t} - \Delta_{i,t} \right\|_2^1  ).
\end{equation}

The $Loss_{\text{proj}}$ can be used to train both length prediction network and direction prediction network using estimated joint position at different stages. With the estimated joint position in the bone length prediction stage, $Loss_{\text{proj-len}}$ is computed to train the length prediction network. $ Loss_{\text{proj-dir}} $ is calculated by the final estimated joint position to train the direction prediction network. 

As shown in Figure \ref{fig:proj benefit}, (a) and (b) have the same single-frame projection loss. However,  the projection consistency loss of these two situations are very different. Therefore, on the one hand, adding the projection consistency loss can constrain the network and make estimated joint position more smooth between frames. On the other hand, by constraining the displacement of the joints, the positions of the joints in adjacent frames can be coupled. The joint position of each frame can be constrained in the motion direction of joints from the previous frame to improve the accuracy of the single-frame joint position.

\begin{table}[t]
	\centering
	\resizebox{1.00\columnwidth}{!}
	{
		\begin{tabular}{|c|c|c|c|c|}
			\hline
			& Protocol 1 & Protocol 2 & Protocol 3 & MPJVE \\ \hline
			Martinez et al. \cite{Martinez_2017_ICCV}  & 45.5  & 37.1 & - & - \\
			Hossain \& Little \cite{hossain2018exploiting} & 41.6 & 31.7 & -  & - \\
			Lee et al. \cite{2018Propagating} &38.4 & - & - & - \\
			Pavllo et al.  \cite{20183D} (243 frames) &37.2  & 27.2 & - & - \\ \hline 
			Chen et al.  \cite{2020Anatomy} (9 frames)   & 38.0       & 28.2       & 37.3       & 1.96     \\
			Ours (9 frames)          & 35.4       & 27.2       & 34.7       & 1.92     \\ \hline
			Chen et al.  \cite{2020Anatomy} (243 frames) & 34.0       & 25.9       & 33.3       & 1.71     \\
			Ours (243 frames)        & 32.5       & 25.2       & 31.9       & 1.70     \\ \hline
		\end{tabular}
	}
	\vspace{-0pt}
	\caption{The results of 9-frame and 243-frame model on Human3.6M with the ground truth 2D input.}
	\label{table:gt}
	\vspace{-0pt}
\end{table}

\subsection{Other Loss Functions}
\label{subsection:3.4}
The loss function of the bone length prediction network is:
\begin{equation}
	Loss_{\text{length}} = \frac{1}{\left| \mathbb J \right|}\sum_{j \in \mathbb J}\left\| P_j - \hat P_{j,L} \right\|_2,
\end{equation}
where $\mathbb J$ represents the set of all joints. $\hat P_{j,L}$ represents the 3D position of the $j$-th joint estimated by bone length estimated network. $P_j$ represents the ground truth of 3D position of the joint. \par 
The loss function of the bone length attention network is:
$Loss_{\text{att}} = \left\| L - \hat L \right\|_2$,
where $\hat L$ represents the length of bones estimated by attention network. $L$ represents the ground truth of bone lengths.

\begin{table}[]
	\begin{center}
		\resizebox{0.9\columnwidth}{!}
		{
			\begin{tabular}{|cc|cccc|}
				\hline
				$w_{pl}$ & $w_{pd}$ & \small{Protocol 1}        & \small{Protocol 2}       & \small{Protocol 3}       & \small{MPJVE}         \\ \hline
				0              & 0.01           & 36.6          & 27.9          & 35.7          & 1.94          \\
				0              & 0.1            & 36.6          & 27.9          & 35.9          & 1.93          \\
				0              & 1              & 35.9 & 27.6 & 35.5 & 1.94 \\
				0              & 10             & 36.5          & 27.9          & 36            & 1.95          \\ \hline
				0.01           & 0              & 36.5          & 27.8          & 36            & 1.95          \\
				0.1            & 0              & 37            & 28.5          & 36.3          & 1.95          \\
				1              & 0              & 36.2 & 27.7 & 35.5 & 1.93 \\
				10             & 0              & 37.2          & 28.4          & 36.1          & 1.96          \\ \hline
			\end{tabular}
		}
	\end{center}
	\vspace{-0pt}
	\caption{Parameter sensitivity test of 9-frame model on Human3.6M dataset, using ground truth 2D joint positions as inputs. In the table, $w_{pl}$ means $w_{proj-len}$ and $w_{pd}$ means $w_{proj-dir}$.}
	\label{tab:parameter}
	\vspace{-0pt}
\end{table}

\begin{figure*}
	\centering
	
	\begin{minipage}[t]{\figheight\textwidth}
		\begin{minipage}[t]{0.23\textwidth}
			\centering
			\includegraphics[width=\linewidth, height = \linewidth]{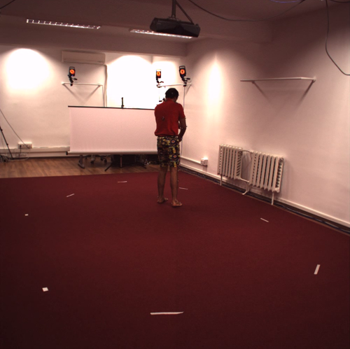}
			\caption*{Image}
		\end{minipage}
		\begin{minipage}[t]{0.23\textwidth}
			\centering
			\includegraphics[width=\linewidth, height = \linewidth]{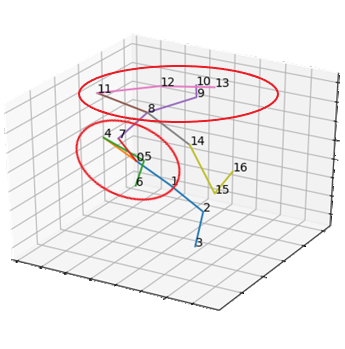}
			\caption*{Baseline}
		\end{minipage}
		\begin{minipage}[t]{0.23\textwidth}
			\centering
			\includegraphics[width=\linewidth, height = \linewidth]{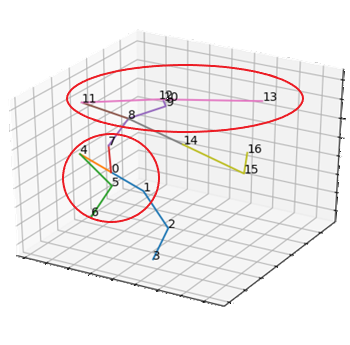}
			\caption*{Ours}
		\end{minipage}
		\begin{minipage}[t]{0.23\textwidth}
			\centering
			\includegraphics[width=\linewidth, height = \linewidth]{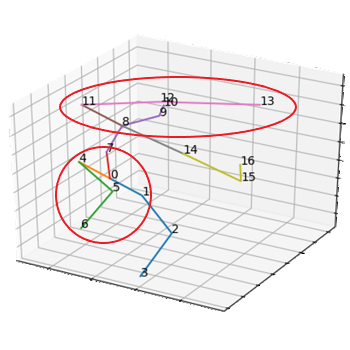}
			\caption*{Ground truth}
		\end{minipage}
	\end{minipage}
	\begin{minipage}[t]{\figheight\textwidth}
		\begin{minipage}[t]{0.23\textwidth}
			\centering
			\includegraphics[width=\linewidth, height = \linewidth]{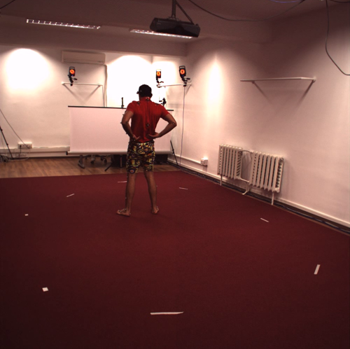}
			\caption*{Image}
		\end{minipage}
		\begin{minipage}[t]{0.23\textwidth}
			\centering
			\includegraphics[width=\linewidth, height = \linewidth]{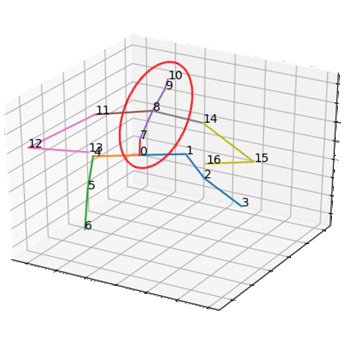}
			\caption*{Baseline}
		\end{minipage}
		\begin{minipage}[t]{0.23\textwidth}
			\centering
			\includegraphics[width=\linewidth, height = \linewidth]{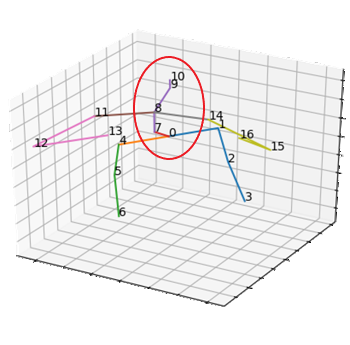}
			\caption*{Ours}
		\end{minipage}
		\begin{minipage}[t]{0.23\textwidth}
			\centering
			\includegraphics[width=\linewidth, height = \linewidth]{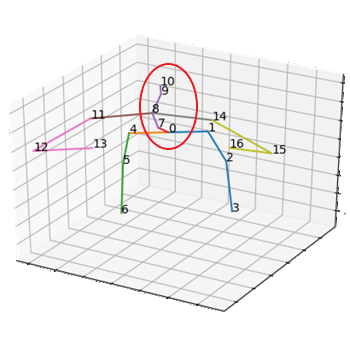}
			\caption*{Ground truth}
		\end{minipage}
	\end{minipage}
	\begin{minipage}[t]{\figheight\textwidth}
		\begin{minipage}[t]{0.23\textwidth}
			\centering
			\includegraphics[width=\linewidth, height = \linewidth]{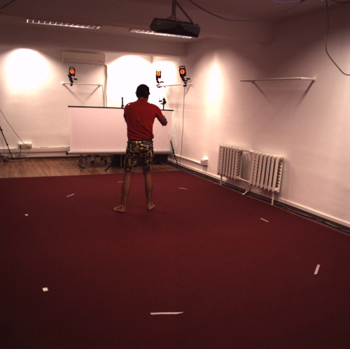}
			\caption*{Image}
		\end{minipage}
		\begin{minipage}[t]{0.23\textwidth}
			\centering
			\includegraphics[width=\linewidth, height = \linewidth]{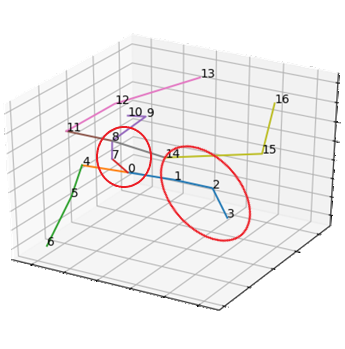}
			\caption*{Baseline}
		\end{minipage}
		\begin{minipage}[t]{0.23\textwidth}
			\centering
			\includegraphics[width=\linewidth, height = \linewidth]{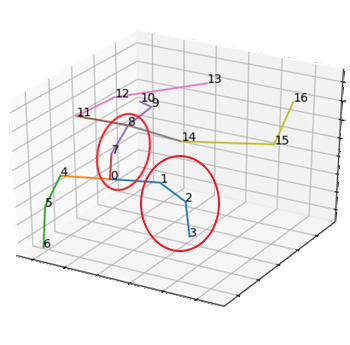}
			\caption*{Ours}
		\end{minipage}
		\begin{minipage}[t]{0.23\textwidth}
			\centering
			\includegraphics[width=\linewidth, height = \linewidth]{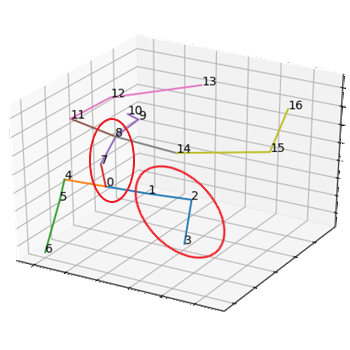}
			\caption*{Ground truth}
		\end{minipage}
	\end{minipage}
	\begin{minipage}[t]{\figheight\textwidth}
		\begin{minipage}[t]{0.23\textwidth}
			\centering
			\includegraphics[width=\linewidth, height = \linewidth]{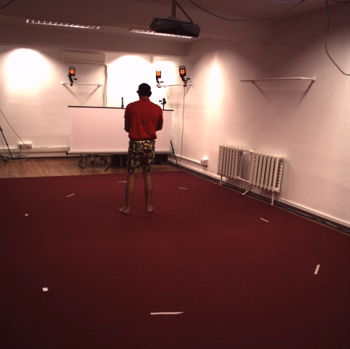}
			\caption*{Image}
		\end{minipage}
		\begin{minipage}[t]{0.23\textwidth}
			\centering
			\includegraphics[width=\linewidth, height = \linewidth]{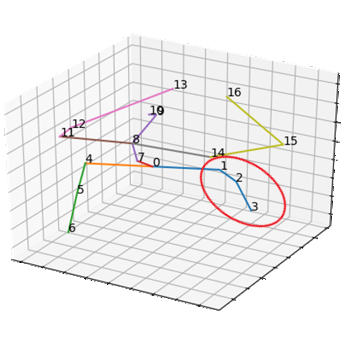}
			\caption*{Baseline}
		\end{minipage}
		\begin{minipage}[t]{0.23\textwidth}
			\centering
			\includegraphics[width=\linewidth, height = \linewidth]{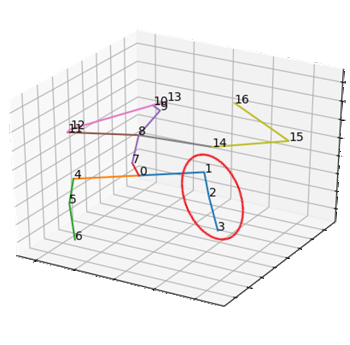}
			\caption*{Ours}
		\end{minipage}
		\begin{minipage}[t]{0.23\textwidth}
			\centering
			\includegraphics[width=\linewidth, height = \linewidth]{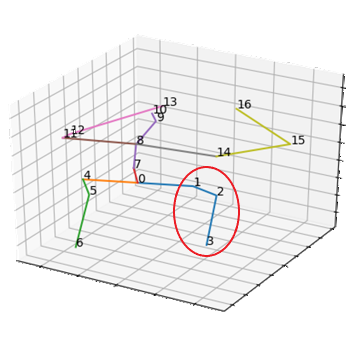}
			\caption*{Ground truth}
		\end{minipage}
	\end{minipage}

	\vspace{-0pt}
	\caption{Visualized comparison of the results of the baseline and the proposed method, both of which use 9-frame receptive field and are trained on the CPN \cite{20183D} 2D inputs. As shown in the red mark in the picture, our prediction has higher accuracy in the end joints due to the application of virtual bones. In addition, our method also performs better under complex human pose and large motions due to the usage of motion constraints based on projection consistency loss.}
	\label{compare}
	\vspace{-0pt}
\end{figure*}

The loss function of the bone direction prediction network is:
$
Loss_{\text{direction}} = \left\| D - \hat D \right\|_2
$,
where $\hat D$ represents the direction of bones estimated by the direction prediction network. $D$ represents the ground truth of bone directions. \par
The joint shift loss \cite{2020Anatomy} is calculated as follows:
\begin{equation}
	Loss_{\text{js}} = \sum_{i\in \mathbb J,j\in \mathbb J, i\neq j} 
	\left\| \left( P_i - P_j \right) - \left( \hat P_i - \hat P_j \right) \right\|
	N(i,j),
\end{equation}
\begin{equation}
	N(i,j) = \begin{cases}
		1, & i, j \text{ are not connected by a real bone,}\\
		0, &i, j \text{ are connected by a real bone},
	\end{cases}
\end{equation}
where $\hat P_i$ represents the position of the $i$-th joint estimated by the final network. \par 
The loss function of the fully connected layers for the final joint prediction is:
\begin{equation}
	Loss_{\text{fc}} = \frac{1}{\left| \mathbb J \right|}\sum_{j \in \mathbb J}\left\| P_j - \hat P_{j} \right\|_2,
\end{equation}
where $\hat P_{j} $ represents the 3D position of the $j$-th joint estimated by the final fully connected layers. 

In general, the complete loss function is as follows:
\begin{equation}
	\begin{aligned}
		Loss_{\text{total}} &= w_L Loss_{\text{length}} + w_{att} Loss_{\text{att}} + w_d Loss_{\text{direction}} \\
		&+ w_{js} Loss_{\text{js}} + w_{pd} Loss_{\text{proj-dir}} + w_{pl} Loss_{\text{proj-len}} \\
		&+ w_{fc} Loss_{\text{fc}},
	\end{aligned}
\end{equation}
where $ w_L, w_{att}, w_d, w_{js}, w_{pd}, w_{pl} \text{ and } w_{fc}   $ are all hyper-parameters.

\begin{table*}[t]
	\begin{center}
		\resizebox{0.9\textwidth}{!}
		{
			\begin{tabular}{|c|c|c|c|c|c|c|c|c|c|}
				\hline
				&                   & \multicolumn{4}{c|}{9-frame}                                               & \multicolumn{4}{c|}{243-frame}                                            \\ \cline{3-10} 
				&                   & Protocol 1 & Protocol 2 & Protocol 3 & Velocity & Protocol 1 & Protocol 2 & Protocol 3 & Velocity \\ \hline
				& Baseline & 38.0                & 28.2                & 37.3                & 1.96              & 34.0                & 25.9                & 33.3                & 1.71              \\ \hline
				
				\multirow{4}{*}{(a)} & Baseline+5VB      & 37.1                & 27.9                & 36.0                & 1.94              & 34.1                & 25.6                & 33.4               & 1.70              \\
				& Baseline+10VB     & 37.1                & 27.9                & 36.4                & 1.96              & 33.4                & 25.4                & 32.8                & 1.70              \\
				& Baseline+13VB    & 36.7                & 27.4                & 35.8                & 1.93              & 33.5                & 25.4                & 32.6                & 1.69              \\
				& Baseline+23VB     & 36.1                & 27.4                & 35.3                & 1.94              & 33.5                & 25.3                & 32.7                & 1.69              \\ \hline
				(b) & 
				Baseline+PCL  & 36.8                & 27.9                & 36.0                & 1.94              & 34.1                & 26.2                & 33.6                & 1.70              \\ \hline
			\end{tabular}
		}
	\end{center}
	\vspace{-0pt}
	\caption{Comparison of different models under Protocols on Human3.6M. "Baseline" represents the baseline 9-frame and 243-frame model we experiment based on  \cite{2020Anatomy}. Other rows represent the loss or virtual bones we proposed. They are used respectively to test the impact on the baseline model. "5VB, 10VB, 13VB, 23VB" refer to the different numbers of virtual bones selected to be added to the bone prediction network. "PCL" refers to the projection consistency loss.}
	\label{table:ablation}
	\vspace{-0pt}
\end{table*}

\section{Experiments}
\subsection{Dataset and Evaluation}
The proposed method is evaluated on Human3.6M dataset \cite{Catalin2014Human3}. Human3.6M provides annotated 2D and 3D joint positions of 3.6 million video frames, which contain 4 camera views for 15 different activities of 11 subjects. Following previous works  \cite{2018Propagating,hossain2018exploiting,20183D}, the training dataset is built on five subjects (S1, S5, S6, S7, S8). The test dataset is built on two subjects (S9, S11) with a 17-joint skeleton.

Four protocols are used to evaluate the models: Protocol 1 (MPJPE, the mean per-joint position error) measures the mean Euclidean distance between the predicted and ground-truth joint positions. Protocol 2 (P-MPJPE) is the error between the aligned predicted 3D joints position and the ground truth. Protocol 3 (N-MPJPE) is the error between the estimated joint position and the ground truth at the same scale. Velocity errors (MPJVE), the errors of the derivative of the corresponding predicted 3D pose over time, are used to measure the smoothness of the predictions.

\subsection{Implementation Details}
\label{suction:parameters}
The proposed method is tested on Human3.6M dataset, using the 2D coordinates of Cascaded Pyramid Network  \cite{20183D} (CPN) or the 2D coordinates of ground truth as the model input. The visibility score used in the proposed method comes from AlphaPose \cite{fang2017rmpe}. The results of the baseline method \cite{2020Anatomy} are experimented based on their open source codes.

The optimizer of the network is Adam~\cite{kingma2014adam}. {The batchsize $b$ is 2048 for the 9-frame model (9-frame receptive field) and 1024 for the 243-frame model (243-frame receptive field). The  number of training epochs are 60. Learning rate is set to $0.001$ and the learning rate decays at the rate of 0.95 per epoch.} We set $ w_{proj-dir} = 1, w_{proj-len} = 1$ for the total loss function. The other hyper-parameters are the same as \cite{2020Anatomy}, such as $w_d = 0.02, w_l = 1, w_{att} = 0.05, w_{js} = 0.1 \text{ and } w_{fc} = 1$. In addition, because the bone direction prediction network can only give the bone direction estimation for the middle frame for each video, the middle frames of different videos under the same camera are used to replace the adjacent frames to calculate $Loss_{\text{proj-dir}}$.
Three NVIDIA 1080Ti GPUs are used to train the 243-frame model and one for the 9-frame model.

\subsection{Experiment Results}
Table \ref{table:ex} and Table \ref{table:gt} show the quantitative comparison of the accuracy between the proposed method and other existing methods on the Human3.6M dataset. For using CPN~\cite{20183D} 2D inputs in Table \ref{table:ex}, our method achieves performance similar to state-of-the-art methods when using a large model with the 243-frame receptive fields. However, for using 2D truth values as input, our method outperforms the state-of-the-art method\cite{2020Anatomy} on all estimation protocols. Therefore, our slightly worse performance when using CPN~\cite{20183D} 2D inputs is due to errors of the input information. When using accurate information as input, that is, 2D joint ground truth, our method shows high performance.
Under the experimental condition of 9-frame receptive fields, the proposed method gets better results than \cite{2020Anatomy} both with CPN~\cite{20183D} 2D inputs and the 2D ground truth inputs. 
The good performance of the proposed method using a smaller model means less computational resources and time consumption. 
Figure \ref{compare} is the visualized results of the proposed method in the two actions, phoning and walking-dog.

Table \ref{tab:parameter} shows the parameter sensitivity of the proposed method. Only the parameters we introduced are tested. The proposed method is sensitive to the choice of super-parameters, so their values are set based on the test results in the table.

{In addition, we also tested the time statistics of the proposed method. In our method, we have a lightweight model (9-frame) and a highly accurate model (243-frame). In recent years, most videos for human pose estimation are recorded at a frequency of 25Hz, so in the real-time test, we believe that the prediction time of 25 frames is less than 1s to meet the real-time performance. For the 9-frame model, the time to predict 25 frames is 0.54s, less than 1s. For the 243-frame model, the time is 1.19s, more than 1s. The lightweight model meets real-time requirement.}

\subsection{Ablation Study}
\vspace{-0pt}
\label{subsec:ablation}
The ablation experiments are performed on Human3.6M under Protocol 1, 2, 3, and Velocity. The 9-frame models and 243-frame models are used respectively for the comparisons between the baseline \cite{2020Anatomy} and the proposed method. Except for the conditions to be compared, other experimental settings are the same as Section \ref{suction:parameters}.

\subsubsection{Influence of Numbers of Virtual Bones} According to the comparison of the results in Table \ref{table:ablation}(a), when 5 virtual bones are added, most protocols are slightly improved. Too few virtual bones added to the joint prediction can only bring a slight improvement. When 10, 13, 23 virtual bones are added, all evaluation protocols have great improvement compared to the baseline. In a smaller receptive field,  the proposed method has produced better results. Our method can have a greater improvement when there is less information input, which shows that our method is more effective when there is more room for improvement.

\subsubsection{Influence of Projection Consistency Loss} It can be seen from the Table \ref{table:ablation}(b) that although increasing the projection consistency loss alone has little effect on the 243-frame model, it can effectively improve the performance of baseline under the 9-frame receptive field.
Another point worth paying attention to is that increasing the projection consistency loss alone has a greater improvement on Protocol 1 and Protocol 3, but the impact on Protocol 2 is more limited. 
Protocol 2 is obtained by calculating the minimum error of the skeleton after the rigid body transformation, so the accuracy of the joint position after rotation is less improved.
\section{Conclusion}
\vspace{-0pt}
In this paper, a novel 3D human pose prediction network and a novel projection consistency loss are proposed. Virtual bones between non-adjacent joints are proposed to optimize the estimation of bone length. Random frames are used to predict the real bone length combining with an attention mechanism, and the current frame is used to predict virtual bone length directly. The bone direction prediction network is implemented by a temporal convolutional network to predict direction. Moreover, a 2D projection consistency loss is presented to constrain the motion displacement of joints between adjacent frames. Experiments indicated that the improved framework performs well in 9-frame receptive field. The study of graphs composed of real bones and virtual bones based on graph networks will be our future direction.

\ifCLASSOPTIONcaptionsoff
  \newpage
\fi

\bibliographystyle{IEEEtran}  
\bibliography{IEEEabrv,bare_jrnl} 

\begin{IEEEbiography}[{\includegraphics[width=1in,height=1.25in,clip,keepaspectratio]{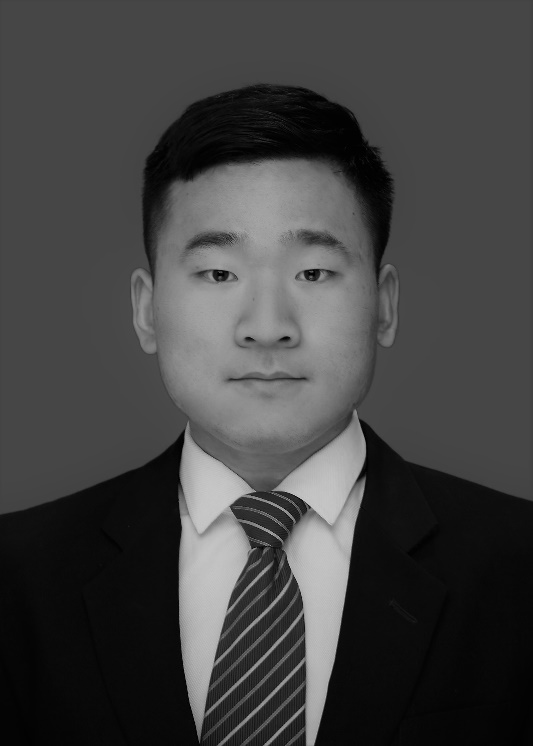}}]{Guangming Wang} (Graduate Student Member,
	IEEE) received the B.S. degree from Department of Automation from Central South University, Changsha, China, in 2018. He is currently pursuing the Ph.D. degree in Control Science and Engineering with Shanghai Jiao Tong University. His current research interests include computer vision and SLAM, in particular, 3D human pose estimation.
\end{IEEEbiography}
\begin{IEEEbiography}[{\includegraphics[width=1in,height=1.25in,clip,keepaspectratio]{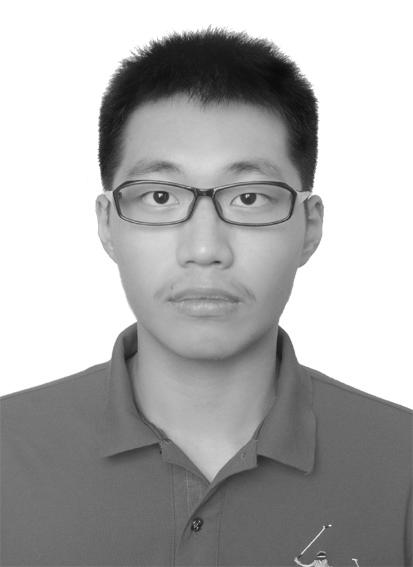}}]{Honghao Zeng}
is currently pursuing the B.S. degree with the Department of Automation, Shanghai Jiao Tong University. His current research interests include computer vision and SLAM, in particular, 3D human pose estimation.
\end{IEEEbiography}
\begin{IEEEbiography}[{\includegraphics[width=1in,height=1.25in,clip,keepaspectratio]{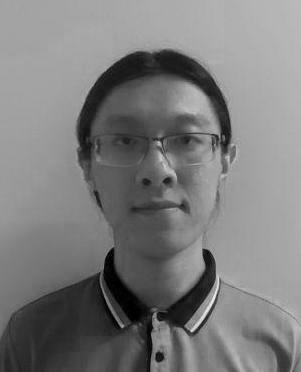}}]{Ziliang Wang}
is currently pursuing the B.S. degree with the Department of Automation, Shanghai Jiao Tong University. His current research interests include computer vision and SLAM, in particular, 3D human pose estimation.
\end{IEEEbiography}

\begin{IEEEbiography}[{\includegraphics[width=1in,height=1.25in,clip,keepaspectratio]{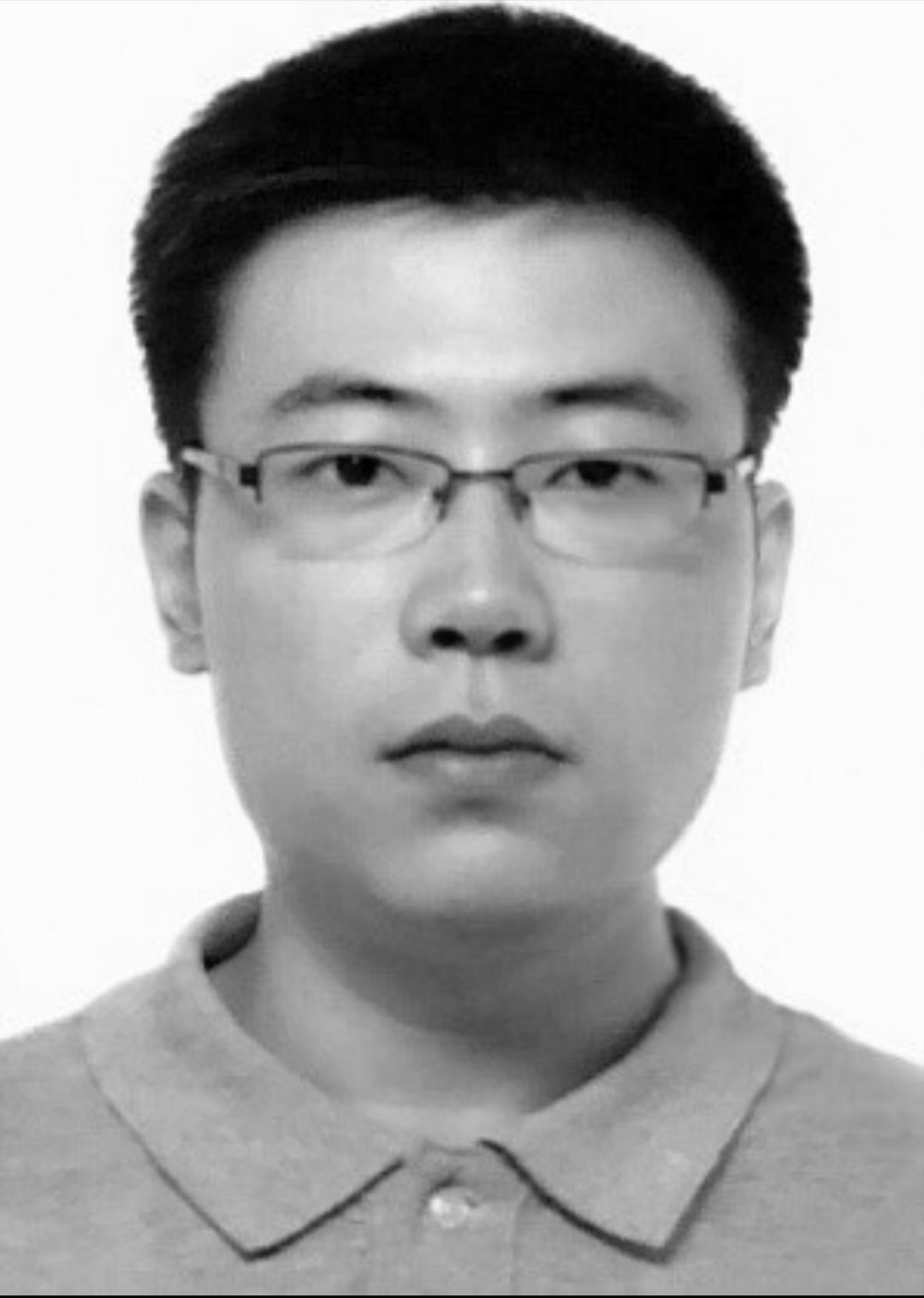}}]{Zhe Liu} received his B.S. degree in Automation from Tianjin University, Tianjin, China, in 2010, and Ph.D. degree in Control Technology and Control Engineering from Shanghai Jiao Tong University, Shanghai, China, in 2016. From 2017 to 2020, he was a Post-Doctoral Fellow with the Department of Mechanical and Automation Engineering, The Chinese University of Hong Kong, Hong Kong. He is currently a Research Associate with the Department of Computer Science and Technology, University of Cambridge. His research interests include autonomous mobile robot, multirobot cooperation and autonomous driving system. 
\end{IEEEbiography}
\begin{IEEEbiography}[{\includegraphics[width=1in,height=1.25in,clip,keepaspectratio]{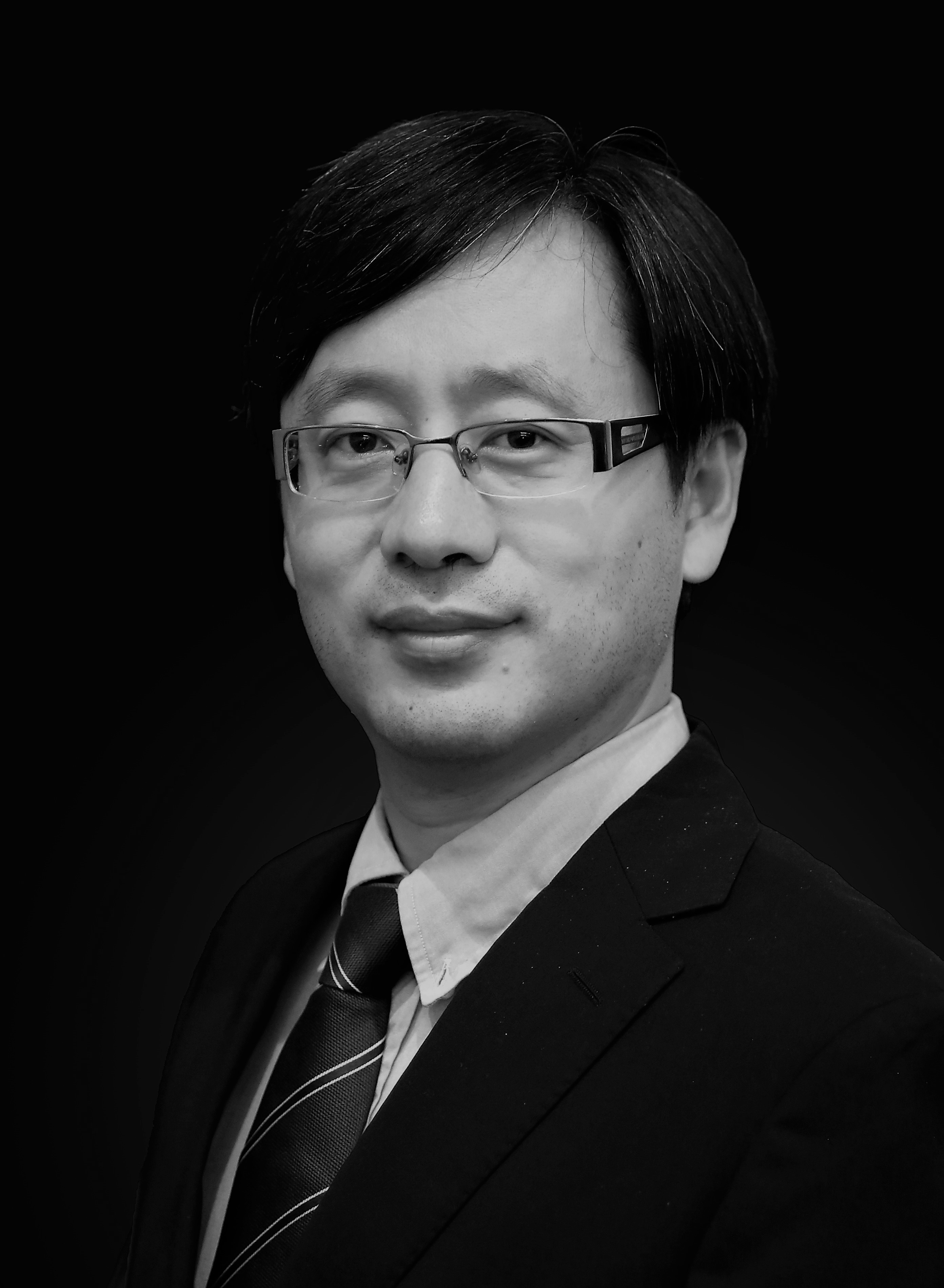}}]{Hesheng Wang}
 (Senior Member, IEEE) received the B.Eng. degree in electrical engineering from the Harbin Institute of Technology, Harbin, China, in 2002, and the M.Phil. and Ph.D. degrees in automation and computer-aided engineering from The Chinese University of Hong Kong, Hong Kong, in 2004 and 2007, respectively. He is currently a Professor with the Department of Automation, Shanghai Jiao Tong University, Shanghai, China. His current research interests include visual servoing, service robot, computer vision, and autonomous driving. 

Dr. Wang is an Associate Editor of IEEE Transactions on Automation Science and Engineering, IEEE Robotics and Automation Letters, Assembly Automation and the International Journal of Humanoid Robotics, a Technical Editor of the IEEE/ASME Transactions on Mechatronics, an Editor of Conference Editorial Board of IEEE Robotics and Automation Society. He served as an Associate Editor of the IEEE Transactions on Robotics from 2015 to 2019. He was the General Chair of IEEE ROBIO 2022 and IEEE RCAR 2016, and the Program Chair of the IEEE ROBIO 2014 and IEEE/ASME AIM 2019. He will be the General Chair of IEEE/RSJ IROS 2025. 
\end{IEEEbiography}




\end{document}